\documentclass[11pt,a4paper]{article}
\usepackage[hyperref]{acl2018}
\usepackage{times}
\usepackage{latexsym}

\usepackage{url}
\usepackage{booktabs} 
\usepackage{color}
\usepackage{latexsym}
\usepackage{amsmath}
\usepackage{amssymb}
\usepackage{enumitem}
\usepackage{listings}
\usepackage{multirow}
\usepackage{tikz}
\usepackage{subcaption}
\usepackage{wrapfig}
\usepackage{graphicx}
\usepackage{soul}
\usepackage[framemethod=TikZ]{mdframed}

\newcommand{\ignore}[1]{}

\aclfinalcopy 

\title{Dynamic Integration of Background Knowledge in Neural NLU Systems}

\author{Dirk Weissenborn \\
German Research Center for AI \\
{\tt dirk.weissenborn@dfki.de}
\And
{Tom\'{a}\v{s} Ko\v{c}isk\'{y} \& Chris Dyer} \\
DeepMind \\
{\tt \{tkocisky,cdyer\}@google.com}
}

\date{}

\begin{document}
\maketitle
\begin{abstract}
Common-sense and background knowledge is required to understand natural language, but in most neural natural language understanding (NLU) systems, this knowledge must be acquired from training corpora during learning, and then it is static at test time. We introduce a new architecture for the dynamic integration of explicit background knowledge in NLU models. A general-purpose reading module reads background knowledge in the form of free-text statements (together with task-specific text inputs) and yields refined word representations to a task-specific NLU architecture that reprocesses the task inputs with these representations. Experiments on document question answering (DQA) and recognizing textual entailment (RTE) demonstrate the effectiveness and flexibility of the approach. Analysis shows that our model learns to exploit knowledge in a semantically appropriate way.
\end{abstract}

\section{Introduction}
Understanding natural language depends crucially on common-sense and background knowledge, for example, knowledge about what concepts are expressed by the words being read (lexical knowledge), and what relations hold between these concepts (relational knowledge). As a simple illustration, if an agent needs to understand that the statement ``King Farouk signed his abdication" is entailed by ``King Farouk was exiled to France in 1952, after signing his resignation", it must know (among other things) that \emph{abdication} means \emph{resignation of a king}.

In most neural natural language understanding (NLU) systems, the requisite background knowledge is implicitly encoded in the models' parameters. That is, what background knowledge is present has been learned from task supervision and also by pre-training word embeddings (where distributional properties correlate with certain kinds of useful background knowledge, such as semantic relatedness). However, acquisition of background knowledge from static training corpora is limiting for two reasons. First, it is unreasonable to expect that all background knowledge that could be important for solving an NLU task can be extracted from a limited amount of training data. Second, as the world changes, the facts that may influence how a text is understood will likewise change. In short: building suitably large corpora to capture all relevant information, and keeping the corpus and derived models up to date with changes to the world would be impractical.

In this paper, we develop a new architecture for dynamically incorporating external background knowledge in NLU models. Rather than relying only on static knowledge implicitly present in the training data, supplementary knowledge is retrieved from external knowledge sources (in this paper, ConceptNet and Wikipedia) to assist with understanding text inputs. Since NLU systems must already read and understand text inputs, we assume that background knowledge will likewise be provided in text form~(\S\ref{sec:external_knowledge}).
The retrieved supplementary texts are read together with the task inputs by an initial reading module whose outputs are \textbf{contextually refined word embeddings}~(\S\ref{sec:reading_stage}). These refined embeddings are then used as input to a task-specific NLU architecture (any architecture that reads text as a sequence of word embeddings can be used here). The initial reading module and the task module are learnt jointly, end-to-end.


We experiment with several different datasets on the tasks of document question answering (DQA) and recognizing textual entailment (RTE) evaluating the impact of our proposed solution with both basic task architectures and a sophisticated task architecture for RTE~(\S\ref{sec:setup}). We find that our embedding refinement strategy is effective~(\S\ref{sec:results}). On four competitive benchmarks, we show that refinement helps. First, simply refining the embeddings just using the context (and no additional background information) can improve performance significantly, but adding background knowledge helps further. Our results are competitive with the best systems, achieving a new state of the art on the recent TriviaQA benchmarks. Our success on this task is especially noteworthy because the task-specific architecture is a simple reading architecture, in particular a single layer BiLSTM with a feed-forward neural network for span prediction. Finally, we provide an analysis demonstrating that our systems are able to exploit background knowledge in a semantically appropriate manner~(\S\ref{sec:analysis}). It includes, for instance, an experiment showing that our system is capable of making appropriate counterfactual inferences when provided with ``alternative facts''.

\section{External Knowledge as Supplementary Text Inputs}\label{sec:external_knowledge}

Knowledge resources make information that could potentially be useful for improving NLU available in a variety different formats, such as natural language text, (subject, predicate, object)-triples, relational databases, and other structured formats. Rather than tailoring our solution to a particular structured representation, we assume that all supplementary information either already exists in natural language statements (e.g., encyclopedias) or can easily be recoded as natural language. Furthermore, while mapping from unstructured to structured representations is hard, the inverse problem is easy. For example, given a triple $(\textit{abdication}, \textsc{isA}, \textit{resignation})$ we can construct the free-text assertion ``Abdication is a resignation.'' using simple rules. Finally, the free-text format means that knowledge that exists only in unstructured text form such as encyclopedic knowledge (e.g., Wikipedia) is usable by our system.

An important question that remains to be answered is: given some text that is to be understood, what supplementary knowledge should be incorporated? The retrieval and preparation of contextually relevant information from knowledge sources is a complex research topic by itself, and there are several statistical \citep{Manning:2008} and more recently neural approaches \citep{mitra2017neural} as well as approaches based on reinforcement learning~\citep{nogueira2017}. Rather than learning both how to incorporate relevant information and which information is relevant, we use a heuristic retrieval mechanism (\S\ref{sec:setup}) and focus on the integration model.

In the next section, we turn to the question of how to leverage the retrieved supplementary knowledge (encoded as text) in a NLU system.

\section{Refining Word Embeddings by Reading}\label{sec:reading_stage}

\begin{figure*}[ht!]
    \centering
    \includegraphics[width=0.8\textwidth]{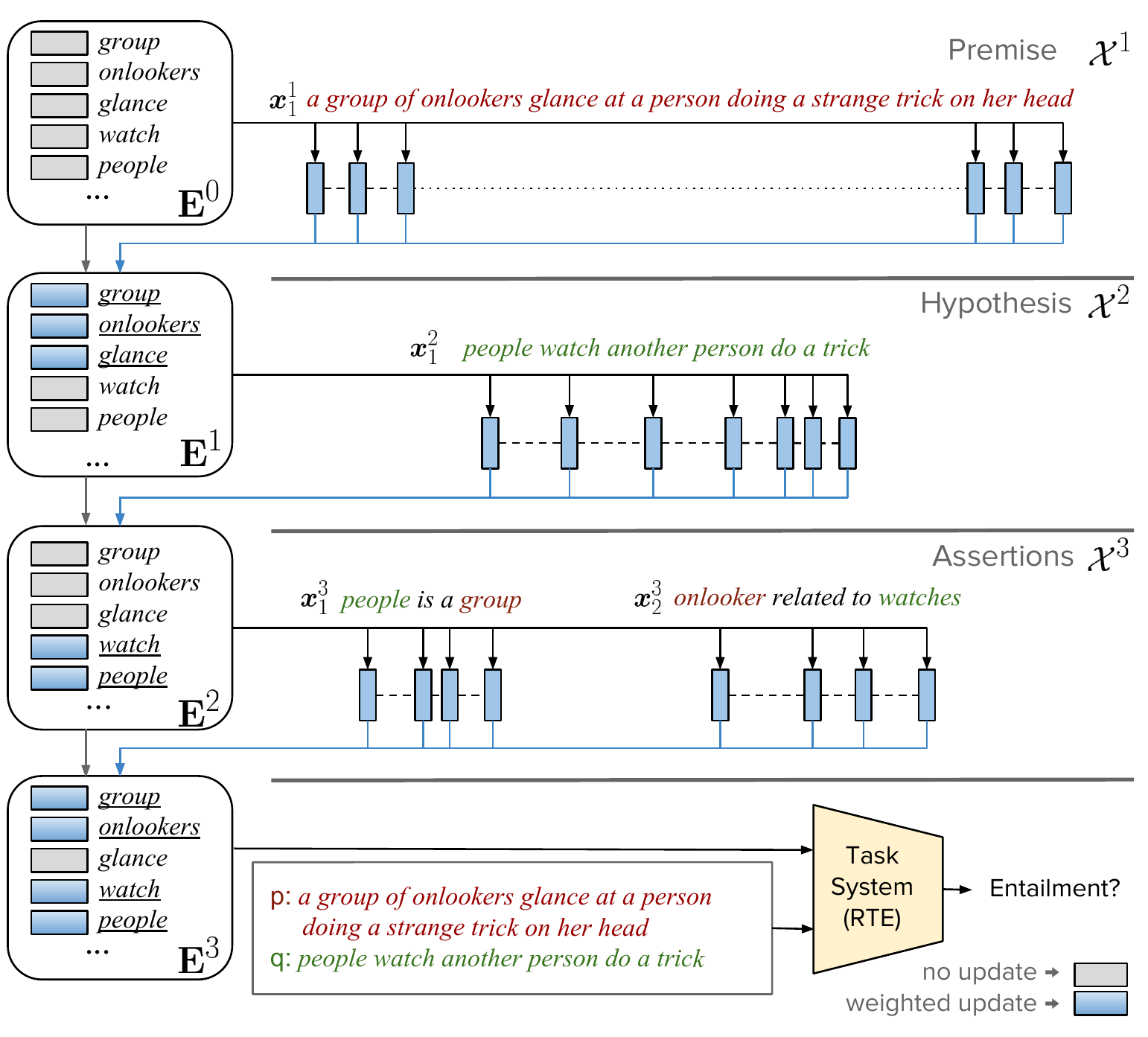}
    \caption{Illustration of our context-dependent, refinement strategy for word representations on an example from the SNLI dataset comprising the premise ($\mathcal{X}_1 = \{\boldsymbol{p}\}$), hypothesis ($\mathcal{X}_2 = \{\boldsymbol{q}\}$) and additional external information in form of free-text assertions from ConceptNet ($\mathcal{X}_1 = \mathcal{A}$). Note that for the QA task there would be another stage that additionally integrates Wikipedia abstracts of answer candidates ($\mathcal{X}_4 = \mathcal{W}$, see \S\ref{sec:setup}). The reading architecture constructs refinements of word representations incrementally (conceptually represented as columns in a series of embedding matrices) $\mathbf{E}^{\ell}$ are incrementally refined by reading the input text and textual renderings of relevant background knowledge before computing the representations used by the task model (in this figure, RTE).}
    \label{fig:illustration}
\end{figure*}

Virtually every NLU task---from document classification to translation to question answering---should in theory be able to benefit from supplementary knowledge. While one could develop custom architectures for each task so as to read supplementary inputs, we would like ours to augment any existing NLU task architectures with the ability to read relevant information with minimal effort. To realize this goal, we adopt the strategy of refining word embeddings; that is, we replace static word embeddings with embeddings that are functions of the task inputs and any supplementary inputs.
Word embeddings can be considered a simple form of key-value memory stores that, in our case, not only contain general-purpose knowledge (as in typical neural NLU systems) but also contextual information (including background knowledge). The use of word-embeddings as memory has the advantage that it is transparent to the task-architecture which kinds of embeddings (refined or unrefined) are used.

Our incremental refinement process \textit{encodes} input texts followed by \textit{updates} on the word embedding matrix in multiple reading steps. Words are first represented non-contextually (i.e., standard word embeddings), which can be conceived of as the columns in an embedding matrix $\mathbf{E}^0$. At each progressive reading step $\ell \ge 1$, a new embedding matrix $\mathbf{E}^\ell$ is constructed by refining the embeddings from the previous step $\mathbf{E}^{\ell-1}$ using (user-specified) contextual information $\mathcal{X}^\ell$ for reading step $\ell$, which is a set of natural language sequences (i.e., texts). 
An illustration of our incremental refinement strategy can be found in Figure~\ref{fig:illustration}.

In the following, we define this procedure formally. We denote the hidden dimensionality of our model by $n$ and a fully-connected layer by $\operatorname{FC}(\mathbf{z})=\mathbf{Wz}+\mathbf{b}$, $\mathbf{W}\in \mathbb{R}^{n \times m}, \mathbf{b}\in\mathbb{R}^n, \mathbf{z}\in\mathbb{R}^m$.

\subsection{Unrefined Word Embeddings ($\mathbf{E}^0$)}\label{sec:noncontextual_embeddings}
The first representation level consists of non-contextual word representations, that is, word representations that do not depend on any input; these can be conceived of as an embedding matrix $\mathbf{E}^0$ whose columns are indexed by words in $\Sigma^*$. The non-contextual word representation $\mathbf{e}^0_w$ for a single word $w$ is computed by using a gated combination of fixed, pre-trained word vectors $\mathbf{e}^p_{w} \in \mathbb{R}^{n^\prime}$ with learned character-based embeddings $\mathbf{e}_w^{\textit{char}} \in \mathbb{R}^n$. We compute $\mathbf{e}_w^{\textit{char}}$ using a single-layer convolutional neural network with $n$ convolutional filters of width $5$ followed by a $\max$-pooling operation over time \citep{Seo2017,Weissenborn2017}. The formal definition of this combination is given in Eq.~\ref{eq:word_embeddings}.

\begin{align}
    \mathbf{e}_w^{p^\prime} &= \operatorname{ReLU}( \operatorname{FC} ( \mathbf{e}_w^p ) ) \nonumber \\
    \mathbf{g}_w &= \operatorname{sigmoid} \left( \operatorname{FC} \begin{bmatrix}  \mathbf{e}_w^{p^\prime} \\ \mathbf{e}_w^{\textit{char}} \end{bmatrix} \right) \nonumber \\
    \mathbf{e}^0_w &= \mathbf{g}_w  \odot \mathbf{e}_w^{p^\prime} + (\mathbf{1}-\mathbf{g}_w) \odot \mathbf{e}_w^{\textit{char}} \label{eq:word_embeddings}
\end{align}

\subsection{Refined Word Embeddings ($\mathbf{E}^{\ell},\, \ell \ge 1$)}
\label{sec:contextual_embeddings}
In order to compute contextually refined word embeddings $\mathbf{E}^{\ell}$ given prior representations $\mathbf{E}^{\ell-1}$ we assume a given set of texts $\mathcal{X}^\ell = \{\boldsymbol{x}^\ell_1, \boldsymbol{x}^\ell_2, \ldots \}$ that are to be read at refinement iteration $\ell$. Each text $\boldsymbol{x}^\ell_i$ is a sequence of word tokens. We embed all tokens of every $\boldsymbol{x}^\ell_i$ using the embedding matrix from the previous layer, $\mathbf{E}^{\ell-1}$. To each word, we concatenate a one-hot vector of length $L$ with position $\ell$ set to $1$, indicating which layer is currently being processed.\footnote{Adding this one-hot feature lets the refinement model learn to update embeddings differently in different levels.} Stacking the vectors into a matrix, we obtain a $\mathbf{X}^\ell_i \in \mathbb{R}^{d \times |\boldsymbol{x}^{\ell}_i|}$. This matrix is processed by a bidirectional recurrent neural network, a $\operatorname{BiLSTM}$ \citep{hochreiter1997long} in this work. The resulting output is further projected to $\hat{\mathbf{X}}^\ell_i$ by a fully-connected layer with $\operatorname{ReLU}$ activation (Eq.~\ref{eq:prepro}).

\begin{align}
    \hat{\mathbf{X}}^\ell_i = \operatorname{ReLU} (\operatorname{FC} (\operatorname{BiLSTM} (\mathbf{X}^\ell_i))) \label{eq:prepro}   
\end{align}

To finally update the previous embedding $\mathbf{e}_w^{\ell-1}$ of word $w$, we initially $\operatorname{maxpool}$ all representations of occurrences matching the lemma of $w$ in every $\boldsymbol{x} \in \mathcal{X}^\ell$ resulting in $\hat{\mathbf{e}}^\ell_w$ (Eq.~\ref{eq:cand_w}). Finally, we combine the previous representation $\mathbf{e}_w^{\ell-1}$ with $\hat{\mathbf{e}}^\ell_w$ to form an updated representation $\mathbf{e}_w^\ell$ via a gated addition. This lets the model determine how much to revise the previous embedding with the newly read information (Eq.~\ref{eq:new_w}).

\begin{align}
    \hat{\mathbf{e}}^\ell_w &= \max \left\{ \hat{\mathbf{x}}^{\ell}_k \mid \operatorname{lemma}(x^{\ell}_k) = \operatorname{lemma}(w) \right\}  \label{eq:cand_w} \\
    \mathbf{u}^\ell_{w} &= \operatorname{sigmoid} \left( \operatorname{FC}\left( \begin{bmatrix}  \mathbf{e}^{\ell - 1}_w \\ \hat{\mathbf{e}}^\ell_w \end{bmatrix} \right) \right)  \label{eq:rep_update_gate} \\    
    \mathbf{e}_w^\ell &= \mathbf{u}^\ell_{w} \odot \mathbf{e}^{\ell-1}_w + (\mathbf{1}-\mathbf{u}^\ell_{w}) \odot \hat{\mathbf{e}}^\ell_w \label{eq:new_w}
\end{align}
\noindent
Note that we soften the matching condition for $w$ using lemmatization,\footnote{\url{https://spacy.io} is used for lemmatization.} $\operatorname{lemma}(w)$, during the pooling operation of Eq.~\ref{eq:cand_w} because contextual information about certain words is usually independent of the current word form $w$ they appear in. As a consequence, this minor linguistic pre-processing step allows for additional interaction between tokens of the same lemma. 

Pooling over lemma-occurrences effectively connects different text passages (even across texts) that are otherwise disconnected, mitigating the problems arising from long-distance dependencies. This is reminiscent of the (soft) attention mechanism used in reading comprehension models (e.g., \newcite{Cheng2016,wang2017gated}). However, our setup is more general as it allows for the connection of multiple passages (via pooling) at once and is able to deal with multiple inputs which is necessary to make use of additional input texts such as relevant background knowledge.

\section{Experimental Setup}
\label{sec:setup}

We run experiments on four benchmarks for two standard NLU tasks: recognizing textual entailment (RTE) and document question answering (DQA). In the following we describe our experimental setup.

\paragraph{Task-specific Models}
Since we wish to assess the value of the proposed embedding refinement strategy, we focus on relatively simple task architectures. We use single-layer bidirectional LSTMs (BiLSTMs) as encoders of the inputs represented by the refined or unrefined embeddings with a task-specific, feed-forward network for the final prediction. Such models are general reading architectures \cite{Bowman2015,Rocktschel2015,Weissenborn2017}. To demonstrate that our reading module can be integrated into arbitrary task architectures, we also add our refinement module to a reimplementation of a state of the art architecture for RTE called ESIM \citep{Chen2017_ESIM}. We refer the interested reader to the ESIM paper for details of the model.

All models are trained end-to-end jointly with the refinement module using a dimensionality of $n=300$ for all but the TriviaQA experiments for which we had to reduce $n$ to $150$ due to memory constraints. All baselines operate on the unrefined word embeddings $\mathbf{E}_0$ described in \S\ref{sec:noncontextual_embeddings}. For the DQA baseline system we add the \textit{lemma-in-question} feature (liq) suggested in \citet{Weissenborn2017}. Implementation details for the BiLSTM task architectures, as well as training details, are available in Appendix~\ref{sec:impl}.

\paragraph{Question Answering} We use 2 recent DQA benchmark training and evaluation datasets, SQuAD \citep{Rajpurkar2016} and TriviaQA \citep{JoshiTriviaQA2017}. The task is to predict an answer span within a provided document $\boldsymbol{p}$ given a question $\boldsymbol{q}$. Both datasets are large-scale, containing on the order of 100k examples, however, TriviaQA is more complex in that the supporting documents are much larger than those for SQuAD. Because TriviaQA is collected via distant supervision the test set is divided into a large but noisy distant supervision part and a much smaller (on the order of hundreds) human verified part. We report results on both. See Appendix~\ref{sec:qa_impl} for implementation details of the DQA system.

\paragraph{Recognizing Textual Entailment} We test on both the SNLI dataset \citep{Bowman2015}, a collection of $570k$ sentence pairs, and the more recent MultiNLI dataset ($433k$ sentence pairs) \citep{williams2017broad}. Given two sentences, a premise $\boldsymbol{p}$ and a hypothesis $\boldsymbol{q}$, the task is to determine whether $\boldsymbol{p}$ either \textit{entails}, \textit{contradicts} or is \textit{neutral} to $\boldsymbol{q}$. See Appendix~\ref{sec:rte_impl} for implementation details of the RTE system.

\paragraph{Supplementary Knowledge Sources} We use \textbf{ConceptNet}\footnote{\url{http://conceptnet.io/}} \citep{Speer2012}, a freely-available, multi-lingual semantic network that originated from the Open Mind Common Sense project and incorporates selected knowledge from various other knowledge sources, such as Wiktionary, Open Multilingual WordNet, OpenCyc and DBpedia. It presents information in the form of relational triples.\footnote{We exclude ConceptNet 4 assertions created by only one contributor and from Verbosity to reduce noise.} Additionally, we exploit \textbf{Wikipedia} abstracts in our DQA experiments as described below.

\paragraph{ConceptNet Integration}
Here we describe the heuristic we use to obtain plausibly relevant supplementary knowledge for understanding a text pair $(\boldsymbol{p},\boldsymbol{q})$ from ConceptNet. Our hypothesis is that relations that link words and phrases across $\boldsymbol{p}$ and $\boldsymbol{q}$ are likely to be most valuable. Because assertions $\boldsymbol{a}$ in ConceptNet come in form of (subject, predicate, object)-triples $(s,r,o)$, we retrieve all assertions for which $s$ appears in $\boldsymbol{q}$ and $o$ appears in $\boldsymbol{p}$, or vice versa. Because still too many such assertions might be retrieved for an instance, we rank all retrievals based on their respective subject and object. The ranking score we use is the inverse product of appearances of the subject and the object in the KB, that is $\operatorname{score}(\boldsymbol{a}) = \left( \sum_{\boldsymbol{a}^\prime} \mathbb{I}(s_{a^\prime} = s_a) \cdot \sum_{\boldsymbol{a}^\prime} \mathbb{I}(o_{a^\prime} = o_a) \right)^{-1}$, where $\mathbb{I}$ denotes the indicator function.
During training and evaluation we retain the top-$k$ assertions, using $k=50$ for DQA and $k=20$ for RTE. Note that fewer or even no assertions might be retrieved for a particular instance during training and testing.

\paragraph{Wikipedia Integration}
Here we describe the heuristic we use to obtain plausibly relevant supplementary knowledge from Wikipedia. We wish to use Wikipedia abstracts\footnote{Downloaded from \url{http://wiki.dbpedia.org/downloads-2016-10}} as an additional knowledge source to gather more information about the top answer predictions of our DQA model. To this end, we let the system first predict the top-16 answer spans without any information from Wikipedia. For each answer candidate string, we collect abstracts for their 3 most frequently linked Wikipedia entries.\footnote{Statistics were extracted from the DBpedia Anchor Text dataset (\url{http://downloads.dbpedia.org/2016-10/core-i18n/en/anchor_text_en.ttl.bz2}).} Using more than only the most frequently linked Wikipedia entry for a given answer string, lets us mitigate problems arising from polysemous entity names, although it does mean the refinement model needs to be selective in extracting relevant information. The refinement module additionally reads the initial 50 tokens of each retrieved Wikipedia abstract and computes the final predictions.

\paragraph{Refinement Order}
When employing our embedding-refinement strategy, we first read the document ($\boldsymbol{p}$) followed by the question ($\boldsymbol{q}$) in case of DQA, and the premise ($\boldsymbol{p}$) followed by the hypothesis ($\boldsymbol{q}$) for RTE, that is, $\mathcal{X}^1=\{\boldsymbol{p}\}$ and $\mathcal{X}^2=\{\boldsymbol{q}\}$. Additional knowledge in the form of a set of assertions $\mathcal{A}$ is integrated after reading the task-specific input for both DQA and RTE, that is, $\mathcal{X}^3 = \mathcal{A}$. Finally, for DQA we additionally add Wikipedia abstracts as background knowledge as described previously, that is, $\mathcal{X}^4 = \mathcal{W}$.
In preliminary experiments we found that the final performance is not significantly sensitive to the order of presentation so we decided to fix our order as defined above.

\section{Results}
\label{sec:results}

This section presents results. We provide ablations for a total of 7 task-dataset-model combinations and compare our final results to other works on the most recent benchmark datasets for each task (TriviaQA and MultiNLI), demonstrating that our results are competitive, and in some cases, state of the art, even without sophisticated task architectures.

\subsection{Question Answering}

\begin{table}[t]
    \small
    \centering
        \begin{tabular}{l c c c}
            \toprule
            \textbf{Model} & \textbf{SQuAD} & \textbf{T-Wiki} & \textbf{T-Web} \\
            \midrule 
            BiLSTM ($\ell = 0$) + liq & 75.9 & 62.1 & 65.0 \\
            + $\boldsymbol{p}$ + $\boldsymbol{q}$ ($\mathbf{E}_2$) & 78.6 & 65.5 & 68.7 \\ 
            + $\boldsymbol{p}$ + $\boldsymbol{q}$ + $\mathcal{A}$ ($\mathbf{E}_3$) & \textbf{79.7} & 67.1 & 70.3 \\
            + $\boldsymbol{p}$ + $\boldsymbol{q}$ + $\mathcal{A}$ + $\mathcal{W}$ ($\mathbf{E}_4$) & \textbf{79.7} & \textbf{69.5} & \textbf{72.7} \\
            \bottomrule
        \end{tabular}
        \caption{Ablation on the SQuAD and TriviaQA (T-Wiki and T-Web) development sets for the F1 metric. Information used for embedding refinement: $\boldsymbol{p/q}$- refinement on task input (i.e., document and question); $\mathcal{A}$- top-50 retrieved ConceptNet assertions; $\mathcal{W}$- Wikipedia abstracts for the top-16 answer candidates. The \textit{liq}-feature (lemma-in-question) is only used in the baseline.}\label{tab:qa_results}
\end{table}

\begin{table}[t]
    \small
    \centering
        \begin{tabular}{l l c c}
        \toprule
            \textbf{Dataset} & \textbf{Model} & \textbf{Exact} & \textbf{F1} \\
            \midrule
            TriviaQA Wiki & (1) & 64.0 / 68.0 & 68.9 / 72.9 \\ 
             & Ours & \textbf{64.6} / \textbf{72.8} & \textbf{69.9} / \textbf{77.4} \\
            \midrule
            TriviaQA Web & (1) & 66.4 / \textbf{80.0} & 71.3 / \textbf{83.7} \\ 
             & Ours & \textbf{67.5} / 77.6 &	\textbf{72.8} / 82.0  \\
            \midrule
            SQuAD Dev & (1) & \textbf{71.6} & \textbf{80.8} \\ 
                & Ours & 69.5 & 79.7 \\ 
            \bottomrule
        \end{tabular}
        \caption{Test set results of our full model (BiLSTM+$\boldsymbol{p}$+$\boldsymbol{q}$+$\mathcal{A}$+$\mathcal{W}$, i.e., using $\mathbf{E}_4$ as embeddings). Results for TriviaQA are divided by distant supervision results (left) and human verified results (right, comprise only several hundreds of examples). We compare against the concurrent work on TriviaQA of \newcite{clark2017simple}.}\label{tab:qa_test_results}
\end{table}

Table~\ref{tab:qa_results} presents our results on two question answering benchmarks. The results demonstrate that the introduction of the refinement module helps consistently, and further improvements come from using common sense knowledge from ConceptNet ($\mathcal{A}$). Wikipedia ($\mathcal{W}$) yields further, significant improvements on TriviaQA, slightly outperforming the current state of the art model (Table~\ref{tab:qa_test_results}). This is especially noteworthy given the simplicity of our QA architecture (i.e., a single layer BiLSTM) compared to the previous SotA attained by \newcite{clark2017simple}. The development results on SQuAD\footnote{We do not report test set results for SQuAD due to restrictions on code sharing.} show the same pattern of improvement, but here the results are slightly worse than the model of \newcite{clark2017simple}, and they are way off from the current best-known results (currently at 87\% F1);\footnote{https://rajpurkar.github.io/SQuAD-explorer/} however, our intention with these experiments is to show of the value that external knowledge and our refinement process can bring, not to compete with highly tuned task architectures on a single dataset.

\paragraph{Controlling for computation.} One potential explanation for the improvement obtained using the refinement module is that we are enabling more computation over the information present in the inputs, that is, we are effectively using a deeper architecture. To test whether this might be the case, we also ran an experiment with a 2-layer BiLSTM (+liq). This setup exhibits similar computational complexity and number of parameters to BiLSTM + $\boldsymbol{p}$ + $\boldsymbol{q}$. We found that the second layer did not improve performance, suggesting that pooling over word/lemma occurrences in a given context between layers, is a powerful, yet simple technique.

\subsection{Recognizing Textual Entailment}

\begin{table}[t]
    \small
    \centering
        \begin{tabular}{l c c c}
            \toprule
            \textbf{Model} & \textbf{SNLI} & \textbf{MNLI} & \textbf{MNLI Mis} \\
            \midrule  
            BiLSTM ($\mathbf{E}_0$) & 84.4 & 70.0 & 70.2 \\  
            + $\boldsymbol{p}$ + $\boldsymbol{q}$ ($\mathbf{E}_2$) & 86.1 & 75.3 & 76.3 \\  
            + $\boldsymbol{p}$ + $\boldsymbol{q}$ + $\mathcal{A}$ ($\mathbf{E}_3$) & \underline{86.5} & \underline{76.8} & \underline{77.5} \\
            \midrule
            ESIM ($\mathbf{E}_0$) & \underline{88.2} & 76.8 & 77.3 \\  
            + $\boldsymbol{p}$ + $\boldsymbol{q}$ ($\mathbf{E}_2$) & 88.0 & 77.8 & 78.4 \\   
            + $\boldsymbol{p}$ + $\boldsymbol{q}$ + $\mathcal{A}$ ($\mathbf{E}_3$) & 87.8 & \underline{78.8} & \underline{78.8} \\
            \midrule
            \multicolumn{4}{c}{Test} \\
            \midrule
            ESIM + $\boldsymbol{p}$ + $\boldsymbol{q}$ + $\mathcal{A}$ & 87.3 & 78.2 & 77.0 \\
            SotA & \textbf{88.6}$^1$ & \textbf{78.8}$^2$ & \textbf{77.8}$^2$ \\  
            \bottomrule
        \end{tabular}    
        \caption{Ablation on the SNLI and MultiNLI-Matched and -Mismatched development set and final results on the corresponding test sets. $\boldsymbol{p/q}$- refinement on task input (i.e., premise and hypothesis); $\mathcal{A}$- top-20 retrieved ConceptNet assertions. $^1$\citet{Chen2017_ESIM}, $^2$\citet{gong2017natural}.}\label{tab:rte_results}
\end{table}

Table~\ref{tab:rte_results} shows the results of our RTE experiments. In general, the introduction of our refinement strategy almost always helps, both with and without external knowledge. When providing additional background knowledge from ConceptNet, our BiLSTM based models improve substantially, while the ESIM-based models improve only on the more difficult MultiNLI dataset. Compared to previously published state of the art systems, our models acquit themselves quite well on the MultiNLI benchmark, and competitively on the SNLI benchmark. In parallel to this work, \citet{gong2017natural} developed a novel task-specific architecture for RTE that achieves slightly better performance on MultiNLI than our ESIM + $\boldsymbol{p}$ + $\boldsymbol{q}$ + $\mathcal{A}$ based models.\footnote{Our refinement architecture can be used of course with this new model, but we report ESIM results since that was best when this work was carried out.} It draws attention to the fact that when using our knowledge-enhanced embedding module, on the MultiNLI, the basic BiLSTM task model outperforms the task-specific ESIM model, which is architecturally much more complex and designed specifically for the RTE task.
We do find that there is little impact of using external knowledge on the RTE task with ESIM, although the refinement strategy helps using just $\boldsymbol{p}$ + $\boldsymbol{q}$. A more detailed set of experiments reported in Appendix~\ref{sec:reduction_experiments} shows that by impoverishing the amount of training data and information present in the GloVe embeddings, the positive impact of supplemental information becomes much more pronounced. These results suggest that ESIM is able to learn important background information from the large-scale datasets and from pretrained embeddings, but this can be supplemented when necessary. Nevertheless, both ESIM and our BiLSTM models when trained with knowledge from ConceptNet are sensitive to the semantics of the provided assertions as demonstrated in our analysis in \S\ref{sec:analysis}. We argue that this is a desirable side effect because it makes the predictions of our model more interpretable than those not trained with knowledge. Furthermore, increasing the coverage of assertions in ConceptNet would most likely yield improved performance even without retraining our models.

Finally, we remark that despite careful tuning, our re-implementation of ESIM fails to match the 88\% reported in \citet{Chen2017_ESIM} by 0.8\%; however, with MultiNLI, we find that our implementation of ESIM performs considerably better (by approximately 5\%). The instability of the results suggests, as well as the failure of a custom RTE-architecture to consistently perform well suggests that current SotA RTE models may be overfit to the SNLI dataset.

\subsection{Qualitative Analysis}
\label{sec:analysis}

\begin{table*}[t]
    \small
    \centering
    \begin{tabular}{l p{4.cm} p{2.6cm} p{5.4cm} }
         \toprule
         
         \textbf{p}: &
         His off-the-cuff style \textit{seems} amateurish [...] & 
         the \textit{net} cost of operations. & 
         but uh these guys [...] file their uh their \textit{final} exams [...] \\
         
         \textbf{h}: &
         He didn't \textit{look like} an amateur &
         The \textit{gross} cost. &
         These men filed their \textit{midterm} exams [...] \\
         \midrule
         
         \textbf{a}: &
         look like synonym seem &
         gross antonym net &
         midterm antonym final \\
         
         $\rightarrow$ &
         contradiction &
         contradiction &
         contradiction \\
         \midrule
         
         $\bar{\textbf{a}}$: &
         look like antonym seem &
         gross synonym net &
         midterm synonym final \\
         
         $\rightarrow$ &
         entailment &
         entailment &
         entailment \\
         \bottomrule
    \end{tabular}
    \caption{Three examples for the \textit{antonym} $\leftrightarrow$ \textit{synonym} swapping experiment on MultiNLI. \textbf{p}-premise, \textbf{h}-hypothesis, \textbf{a}-assertion, $\bar{\textbf{a}}$-swapped assertion.}
    \label{tab:examples}
\end{table*}

Although our empirical results show our knowledge-incorporation approach improves performance, in this section we attempt to assess whether we are learning to use the provided knowledge in a semantically appropriate way.

\paragraph{RTE} To test our models sensitivity towards the semantics of the assertions for recognizing textual entailment, we run an experiment in which we swap the \textit{synonym} with the \textit{antonym} predicate in the provided assertions during test time. We hypothesize that in many cases these two predicates are very important for predicting either \textit{contradiction} or \textit{entailment}. Indeed, there is a strong performance drop of about \textbf{10\%} on MultiNLI examples for both the BiLSTM and the ESIM model for which either a \textit{synonym} or an \textit{antonym}-assertion is present. This very large drop clearly shows that our models are sensitive to the semantics of the provided knowledge. Examples of prediction changes are presented in Table~\ref{tab:examples}. They demonstrate that the system has learned to trust the presented assertions to the point that it will make appropriate counterfactual inferences---that is, the change in knowledge has \emph{caused} the change in prediction. For the interested reader we provide additional RTE analysis results in Appendix~\ref{sec:additional_analysis}

\paragraph{DQA} The following is an example question from the TriviaQA dataset:

\begin{mdframed}[roundcorner=2pt]
\small
Prince Philip [\ldots] was born on which island?

\noindent \textit{Answer candidates with corresponding abstracts:}
\begin{itemize}[noitemsep,nolistsep]
\item \textbf{Denmark} is a Scandinavian country with territory in Europe and North America [\ldots]
\item \textbf{Corfu} is a Greek island in the Ionian Sea [\ldots]
\item \textbf{Greece}, officially the Hellenic Republic, [\ldots] is a transcontinental country [\ldots]
\item \textbf{Vanuatu} is a Pacific island nation located in the South Pacific Ocean [\ldots]
\end{itemize}

\end{mdframed}

\noindent Answer candidates (i.e., \textit{Denmark}, \textit{Corfu}, \textit{Greece}, \textit{Vanuata}) were obtained from the top predicted answer spans computed by our model excluding Wikipedia (i.e., BiLSTM + $\boldsymbol{p}$ + $\boldsymbol{q}$ + $\mathcal{A}$). Their corresponding abstracts were retrieved from Wikipedia and then given to our model in a second pass (i.e., BiLSTM + $\boldsymbol{p}$ + $\boldsymbol{q}$ + $\mathcal{A}$ + $\mathcal{W}$). In this example, the final best prediction of the model changes from Denmark to Corfu after integrating the abstracts (here, the abstract clearly states that Corfu is an island). We studied a total of 25 similar answer changes, 14 of which went from incorrect to correct, and 11 of which went from correct to incorrect. In 11 of the 14 corrections, obvious information is present in the Wikipedia abstracts that reinforced the correct answer. Where the system was confused by the answers (i.e., when the abstracts switched the production from correct to incorrect), no obvious information was present in 8 of the 11 cases, suggesting that the model had difficulty coping with unrelated background information. In 3 of the 11, plausibly relevant information was present in the abstract of the correct answer, yet the model still made the incorrect answer change.

The existence of counterfactual inferences in RTE and the tendency to use reinforcing information about candidate answers in DQA suggest that our knowledge incorporating strategy is exploiting heterogeneous knowledge sources in semantically sensible ways.

\section{Related Work}
The role of background knowledge in natural language understanding has long been remarked on, especially in the context of classical models of AI~\citep{schank:1977,minsky2000commonsense}; however, it has only recently begun to play a role in neural network models of NLU \citep{ahn2016neural,Xu2016,long2017world,Dhingra2017}. Previous efforts have focused on specific tasks or certain kinds of knowledge, whereas we take a step towards a more general-purpose solution for the integration of heterogeneous knowledge for NLU systems by providing a simple, general-purpose reading architecture that can read background knowledge encoded in simple natural language statements, e.g., ``abdication is a type of resignation''.

In the area of visual question answering \citet{Wu2016} utilize external knowledge in form of DBpedia comments (short abstracts/definitions) to improve the answering ability of a model. \citet{marino2016more} explicitly incorporate knowledge graphs into an image classification model. \citet{Xu2016} created a recall mechanism into a standard LSTM cell that retrieves pieces of external knowledge encoded by a single representation for a conversation model. Concurrently, \citet{Dhingra2017} exploit linguistic knowledge using MAGE-GRUs, an adapation of GRUs to handle graphs, however, external knowledge has to be present in form of triples. \citet{ahn2016neural} exploit knowledge base facts about mentioned entities for neural language models. \citet{bahdanau2017learning} and  \citet{long2017world} create word embeddings on-the-fly by reading word definitions prior to processing the task at hand. \citet{pilehvar2017towards} incorporate information about word senses into their representations before solving the downstream NLU task, which is similar. We go one step further by seamlessly integrating all kinds of fine-grained assertions about concepts that might be relevant for the task at hand.

Another important aspect of our approach is the notion of dynamically updating word-representations with contextual information. Tracking and updating concepts, entities or sentences with dynamic memories is a very active research direction \citep{kumar2016ask,henaff2017tracking,ji2017dynamic,kobayashi2017neural}. However, those works typically focus on particular tasks whereas our approach is task-agnostic and most importantly allows for the easy integration of external background knowledge. Important progress has also been made in creating pre-trained, contextualized token representations \cite{Peters2017,mccann2017learned}.

\section{Conclusion}

We have presented a novel reading architecture that allows for the dynamic integration of background knowledge into neural NLU models. Our solution, which is based on the incremental refinement of word representations by reading supplementary inputs, is flexible and can be used with virtually any existing NLU architecture that rely on word embeddings as input.  Our results show that embedding refinement using both the system's text inputs, as well as supplementary text from external background knowledge can yield large improvements. In particular, we have shown that relatively simple task architectures (e.g., based on simple BiLSTM readers) can become competitive with state of the art, task-specific architectures when augmented with our reading architecture. Our analysis demonstrates that our model learns to exploit provided background knowledge in a semantically appropriate way.



\bibliography{acl2018}

\begin{thebibliography}{}
\expandafter\ifx\csname natexlab\endcsname\relax\def\natexlab#1{#1}\fi

\bibitem[{Ahn et~al.(2016)Ahn, Choi, P{\"a}rnamaa, and Bengio}]{ahn2016neural}
Sungjin Ahn, Heeyoul Choi, Tanel P{\"a}rnamaa, and Yoshua Bengio. 2016.
\newblock A neural knowledge language model.
\newblock {\em arXiv\/} .

\bibitem[{Bahdanau et~al.(2017)Bahdanau, Bosc, Jastrzebski, Grefenstette,
  Vincent, and Bengio}]{bahdanau2017learning}
Dzmitry Bahdanau, Tom Bosc, Stanislaw Jastrzebski, Edward Grefenstette, Pascal
  Vincent, and Yoshua Bengio. 2017.
\newblock Learning to compute word embeddings on the fly.
\newblock {\em arXiv\/} .

\bibitem[{Bowman et~al.(2015)Bowman, Angeli, {Potts Christopher}, and
  Manning}]{Bowman2015}
Samuel~R Bowman, Gabor Angeli, {Potts Christopher}, and Christopher~D Manning.
  2015.
\newblock {A large annotated corpus for learning natural language inference}.
\newblock In {\em EMNLP\/}.

\bibitem[{Chen et~al.(2017)Chen, Zhu, Ling, Wei, and Jiang}]{Chen2017_ESIM}
Qian Chen, Xiaodan Zhu, Zhenhua Ling, Si~Wei, and Hui Jiang. 2017.
\newblock {Enhancing and Combining Sequential and Tree LSTM for Natural
  Language Inference}.
\newblock {\em ACL\/} .

\bibitem[{Cheng et~al.(2016)Cheng, Dong, and Lapata}]{Cheng2016}
Jianpeng Cheng, Li~Dong, and Mirella Lapata. 2016.
\newblock Long short-term memory-networks for machine reading.
\newblock In {\em EMNLP\/}.

\bibitem[{Clark and Gardner(2017)}]{clark2017simple}
Christopher Clark and Matt Gardner. 2017.
\newblock Simple and effective multi-paragraph reading comprehension.
\newblock {\em arXiv\/} .

\bibitem[{Dhingra et~al.(2017)Dhingra, Yang, Cohen, and
  Salakhutdinov}]{Dhingra2017}
Bhuwan Dhingra, Zhilin Yang, William~W Cohen, and Ruslan Salakhutdinov. 2017.
\newblock {Linguistic Knowledge as Memory for Recurrent Neural Networks}.
\newblock {\em arXiv\/} .

\bibitem[{Gong et~al.(2017)Gong, Luo, and Zhang}]{gong2017natural}
Yichen Gong, Heng Luo, and Jian Zhang. 2017.
\newblock Natural language inference over interaction space.
\newblock {\em arXiv\/} .

\bibitem[{Henaff et~al.(2017)Henaff, Weston, Szlam, Bordes, and
  LeCun}]{henaff2017tracking}
Mikael Henaff, Jason Weston, Arthur Szlam, Antoine Bordes, and Yann LeCun.
  2017.
\newblock Tracking the world state with recurrent entity networks.
\newblock In {\em ICLR\/}.

\bibitem[{Hochreiter and Schmidhuber(1997)}]{hochreiter1997long}
Sepp Hochreiter and J{\"u}rgen Schmidhuber. 1997.
\newblock Long short-term memory.
\newblock {\em Neural computation\/} .

\bibitem[{Ji et~al.(2017)Ji, Tan, Martschat, Choi, and Smith}]{ji2017dynamic}
Yangfeng Ji, Chenhao Tan, Sebastian Martschat, Yejin Choi, and Noah~A Smith.
  2017.
\newblock Dynamic entity representations in neural language models.
\newblock In {\em EMNLP\/}.

\bibitem[{Joshi et~al.(2017)Joshi, Choi, Weld, and
  Zettlemoyer}]{JoshiTriviaQA2017}
Mandar Joshi, Eunsol Choi, Daniel~S. Weld, and Luke Zettlemoyer. 2017.
\newblock Triviaqa: A large scale distantly supervised challenge dataset for
  reading comprehension.
\newblock In {\em ACL\/}.

\bibitem[{Kobayashi et~al.(2017)Kobayashi, Okazaki, and
  Inui}]{kobayashi2017neural}
Sosuke Kobayashi, Naoaki Okazaki, and Kentaro Inui. 2017.
\newblock A neural language model for dynamically representing the meanings of
  unknown words and entities in a discourse.
\newblock {\em arXiv preprint arXiv:1709.01679\/} .

\bibitem[{Kumar et~al.(2016)Kumar, Irsoy, Ondruska, Iyyer, Bradbury, Gulrajani,
  Zhong, Paulus, and Socher}]{kumar2016ask}
Ankit Kumar, Ozan Irsoy, Peter Ondruska, Mohit Iyyer, James Bradbury, Ishaan
  Gulrajani, Victor Zhong, Romain Paulus, and Richard Socher. 2016.
\newblock Ask me anything: Dynamic memory networks for natural language
  processing.
\newblock In {\em ICML\/}. pages 1378--1387.

\bibitem[{Long et~al.(2017)Long, Bengio, Lowe, Cheung, and
  Precup}]{long2017world}
Teng Long, Emmanuel Bengio, Ryan Lowe, Jackie Chi~Kit Cheung, and Doina Precup.
  2017.
\newblock World knowledge for reading comprehension: Rare entity prediction
  with hierarchical lstms using external descriptions.
\newblock In {\em EMNLP\/}.

\bibitem[{Manning et~al.(2008)Manning, Raghavan, and
  Sch\"{u}tze}]{Manning:2008}
Christopher~D. Manning, Prabhakar Raghavan, and Hinrich Sch\"{u}tze. 2008.
\newblock {\em Introduction to Information Retrieval\/}.
\newblock Cambridge University Press, New York, NY, USA.

\bibitem[{Marino et~al.(2017)Marino, Salakhutdinov, and Gupta}]{marino2016more}
Kenneth Marino, Ruslan Salakhutdinov, and Abhinav Gupta. 2017.
\newblock The more you know: Using knowledge graphs for image classification.
\newblock {\em CVPR\/} .

\bibitem[{McCann et~al.(2017)McCann, Bradbury, Xiong, and
  Socher}]{mccann2017learned}
Bryan McCann, James Bradbury, Caiming Xiong, and Richard Socher. 2017.
\newblock Learned in translation: Contextualized word vectors.
\newblock In {\em NIPS\/}.

\bibitem[{Minsky(2000)}]{minsky2000commonsense}
Marvin Minsky. 2000.
\newblock Commonsense-based interfaces.
\newblock {\em Communications of the ACM\/} .

\bibitem[{Mitra and Craswell(2017)}]{mitra2017neural}
Bhaskar Mitra and Nick Craswell. 2017.
\newblock Neural models for information retrieval.
\newblock {\em arXiv preprint arXiv:1705.01509\/} .

\bibitem[{Nogueira and Cho(2017)}]{nogueira2017}
Rodrigo Nogueira and Kyunghyun Cho. 2017.
\newblock Task-oriented query reformulation with reinforcement learning.
\newblock In {\em EMNLP\/}.

\bibitem[{Pennington et~al.(2014)Pennington, Socher, and
  Manning}]{Pennington2014}
Jeffrey Pennington, Richard Socher, and Christopher~D. Manning. 2014.
\newblock Glove: Global vectors for word representation.
\newblock In {\em EMNLP\/}.

\bibitem[{Peters et~al.(2017)Peters, Ammar, Bhagavatula, and
  Power}]{Peters2017}
Matthew~E. Peters, Waleed Ammar, Chandra Bhagavatula, and Russell Power. 2017.
\newblock Semi-supervised sequence tagging with bidirectional language models.
\newblock In {\em ACL\/}.

\bibitem[{Pilehvar et~al.(2017)Pilehvar, Camacho-Collados, Navigli, and
  Collier}]{pilehvar2017towards}
Mohammad~Taher Pilehvar, Jose Camacho-Collados, Roberto Navigli, and Nigel
  Collier. 2017.
\newblock Towards a seamless integration of word senses into downstream nlp
  applications.
\newblock In {\em ACL\/}.

\bibitem[{Rajpurkar et~al.(2016)Rajpurkar, Zhang, Lopyrev, and
  Liang}]{Rajpurkar2016}
Pranav Rajpurkar, Jian Zhang, Konstantin Lopyrev, and Percy Liang. 2016.
\newblock In {\em {SQuAD: 100,000+ Questions for Machine Comprehension of
  Text}\/}.

\bibitem[{Rockt{\"a}schel et~al.(2015)Rockt{\"a}schel, Grefenstette, Hermann,
  Kocisk{\'y}, and Blunsom}]{Rocktschel2015}
Tim Rockt{\"a}schel, Edward Grefenstette, Karl~Moritz Hermann, Tom{\'a}s
  Kocisk{\'y}, and Phil Blunsom. 2015.
\newblock Reasoning about entailment with neural attention.
\newblock {\em ICLR\/} .

\bibitem[{Schank and Abelson(1977)}]{schank:1977}
Roger Schank and Robert Abelson. 1977.
\newblock {\em Scripts, Plans, Goals, and Understanding\/}.
\newblock Psychology Press.

\bibitem[{Seo et~al.(2017)Seo, Kembhavi, Farhadi, and Hajishirzi}]{Seo2017}
Minjoon Seo, Aniruddha Kembhavi, Ali Farhadi, and Hananneh Hajishirzi. 2017.
\newblock {Bi-Directional Attention Flow for Machine Comprehension}.
\newblock In {\em ICLR\/}.

\bibitem[{Speer and Havasi(2012)}]{Speer2012}
Robert Speer and Catherine Havasi. 2012.
\newblock {Representing General Relational Knowledge in ConceptNet 5}.
\newblock In {\em LREC\/}.

\bibitem[{Wang et~al.(2017)Wang, Yang, Wei, Chang, and Zhou}]{wang2017gated}
Wenhui Wang, Nan Yang, Furu Wei, Baobao Chang, and Ming Zhou. 2017.
\newblock Gated self-matching networks for reading comprehension and question
  answering.
\newblock In {\em ACL\/}.

\bibitem[{Weissenborn et~al.(2017)Weissenborn, Wiese, and
  Seiffe}]{Weissenborn2017}
Dirk Weissenborn, Georg Wiese, and Laura Seiffe. 2017.
\newblock {Making Neural QA as Simple as Possible but not Simpler}.
\newblock In {\em CoNLL\/}.

\bibitem[{Williams et~al.(2017)Williams, Nangia, and
  Bowman}]{williams2017broad}
Adina Williams, Nikita Nangia, and Samuel~R Bowman. 2017.
\newblock A broad-coverage challenge corpus for sentence understanding through
  inference.
\newblock {\em arXiv\/} .

\bibitem[{Wu et~al.(2016)Wu, Wang, Shen, van~den Hengel, and Dick}]{Wu2016}
Qi~Wu, Peng Wang, Chunhua Shen, Anton van~den Hengel, and Anthony Dick. 2016.
\newblock {Ask Me Anything: Free-form Visual Question Answering Based on
  Knowledge from External Sources}.
\newblock {\em CVPR\/} .

\bibitem[{Xu et~al.(2016)Xu, Liu, Wang, Sun, and Wang}]{Xu2016}
Zhen Xu, Bingquan Liu, Baoxun Wang, Chengjie Sun, and Xiaolong Wang. 2016.
\newblock {Incorporating Loose-Structured Knowledge into LSTM with Recall Gate
  for Conversation Modeling}.
\newblock {\em arXiv\/} .

\end{thebibliography}
\bibliographystyle{acl_natbib}

\clearpage
\clearpage

\appendix

\section{Implementation Details}\label{sec:impl}

All our models were trained with 3 different random seeds and the top performance is reported \footnote{Result variations were small, that is within less than a percentage point in all experiments.}. An overview of hyper-parameters used in our experiments can be found in Table~\ref{tab:training_details}. In the following we explain the detailed implementation of our two task-specific, baseline models. 

We assume to have computed the contextually (un-)refined word representations depending on the setup and embedded our input sequences $\boldsymbol{q}=(q_1,...,q_{L_Q})$ and  $\boldsymbol{p}=(p_1,...,p_{L_P})$ to $\mathbf{Q} \in \mathbb{R}^{n \times L_Q}$ and $\mathbf{P} \in \mathbb{R}^{n \times L_P}$, respectively. The word representation update gate in Eq.~\ref{eq:rep_update_gate} is initialized with a bias of $1$ to refine representations only slightly in the beginning of training. In the following as before, we denote the hidden dimensionality of our model by $n$ and a fully-connected layer by $\operatorname{FC}(\mathbf{z})=\mathbf{Wz}+\mathbf{b}$, $\mathbf{W}\in \mathbb{R}^{n \times m}, \mathbf{b}\in\mathbb{R}^n, \mathbf{u}\in\mathbb{R}^m$.

\subsection{Question Answering}\label{sec:qa_impl}

\paragraph{Encoding} In the DQA task $\boldsymbol{q}$ refers to the question and $\boldsymbol{p}$ to the supporting text. For our baseline (i.e., BiLSTM + liq) we additionally concatenate a binary feature to $\boldsymbol{p}$ and $\boldsymbol{q}$ indicating whether the corresponding token lemma appeared in the question. However, it is omitted in the following for the sake of brevity. At first we process both sequences by identical $\operatorname{BiLSTM}$s in parallel (Eq.~\ref{eq:qa_bilstm}) followed by a linear projection and a $\tanh$ non-linearity (Eq.~\ref{eq:qa_proj}) .

\begin{align}
    \hat{\mathbf{Q}} &= \operatorname{BiLSTM} \left( \mathbf{Q}\right)
    & \hat{\mathbf{Q}} \in \mathbb{R}^{2n \times L_Q} \nonumber\\
    \hat{\mathbf{P}} &= \operatorname{BiLSTM} \left( \mathbf{P}\right) 
    & \hat{\mathbf{P}} \in \mathbb{R}^{2n \times L_P} \label{eq:qa_bilstm} \\
    \tilde{\mathbf{Q}} &= \tanh \left( \mathbf{U} \hat{\mathbf{Q}} \right)
    & \tilde{\mathbf{Q}} \in \mathbb{R}^{n \times L_Q} \nonumber \\
    \tilde{\mathbf{P}} &= \tanh \left( \mathbf{U} \hat{\mathbf{P}} \right)
    & \tilde{\mathbf{P}} \in \mathbb{R}^{n \times L_P} \label{eq:qa_proj} 
\end{align}
\noindent
$\mathbf{U} \in \mathbb{R}^{n\times 2n}$ is initialized by $[I;I]$ where $I \in \mathbb{R}^{n\times n}$ is the identity matrix.

\paragraph{Prediction} Our prediction-- or answer layer is similar to \newcite{Weissenborn2017}. We first compute a weighted, $n$-dimensional representation $\tilde{\boldsymbol{q}}$ of the processed question $\tilde{\mathbf{Q}}$ (Eq.~\ref{eq:question_representation}).

\begin{align}
    \alpha &= \operatorname{softmax}(\mathbf{v}_q \tilde{\mathbf{Q}}) \quad , \, \mathbf{v}_q \in \mathbb{R}^n \nonumber \\
    \tilde{\mathbf{q}} &= \sum_i \alpha_i \tilde{\mathbf{q}}_i \label{eq:question_representation}
\end{align}

The probability distributions $p_s$/$p_e$ for the start/end location of the answer is computed by a 2-layer MLP with a ReLU activated, hidden layer $\boldsymbol{s}_j$ as follows:

\begin{align}
    \mathbf{s}_j &= \operatorname{ReLU}  \left(\operatorname{FC}_s\left( \begin{bmatrix} \tilde{\mathbf{p}}_j \\ \tilde{\mathbf{q}} \\ \tilde{\mathbf{p}}_j \odot  \tilde{\mathbf{q}} \end{bmatrix} \right) \right) \nonumber\\
    \mathbf{e}_j &= \operatorname{ReLU}  \left(\operatorname{FC}_e\left( \begin{bmatrix} \tilde{\mathbf{p}}_j \\ \tilde{\mathbf{q}} \\ \tilde{\mathbf{p}}_j \odot  \tilde{\mathbf{q}} \end{bmatrix} \right) \right)
    \nonumber \\
    p_{s}(j) &= \operatorname{softmax}_j(\mathbf{v}_s \mathbf{s}_j) \qquad \mathbf{v}_s \in \mathbb{R}^n \nonumber \\
    p_{e}(j) &= \operatorname{softmax}_j(\mathbf{v}_e \mathbf{s}_j) \qquad \mathbf{v}_e \in \mathbb{R}^n
    \label{eq:span_prediction}
\end{align}

The model is trained to maximize the log-likelihood of the correct answer spans by computing the sum of the correct span probabilities  $p_{s}(i) \cdot p_{e}(k)$ for span $(i, k)$ under our model (Eq.~\ref{eq:span_prediction}). During evaluation we extract the span $(i, k)$ with the best score and maximum token length $k-i \le 16$ for SQuAD and $k-i \le 8$ for TriviaQA.

\paragraph{TriviaQA} Properly training a QA system on TriviaQA is much more challenging than SQuAD because of the large document sizes and the use of multiple paragraphs. Therefore, we adopt the approach of \newcite{clark2017simple} who were the first to properly train neural QA models on TriviaQA. It relies on splitting documents and merging paragraphs up to a certain maximum token length ($600$ per paragraph in our experiments), and only retaining the top-$k$ paragraphs ($6$ in our case) for prediction. Paragraphs are ranked using the \textit{tf-idf} cosine similarity between question and paragraph. To speed up training only $2$ paragraphs out of the top $4$/$6$ for the $Web$/$Wikipedia$ datasets were sampled. The only architectural difference for this multi-paragraph setup is that we encode multiple $\boldsymbol{p}$ for each question $\boldsymbol{q}$ and the $\operatorname{softmax}$ of Eq.~\ref{eq:span_prediction} is taken over all tokens of all paragraphs instead of only a single paragraph. For further details, we refer the interested reader to \newcite{clark2017simple} who explain this process in more detail.

\begin{table}[t]
    \small
    \begin{tabular}{l c c c c}
        \toprule
        \textbf{Dataset} & \textbf{Dim} $n$ & \textbf{Drop.} & \textbf{B.-size} & \textbf{Ckpt Interval} \\ \midrule
        SQuAD & 300 & 0.2 & 32 & 1000 \\
        TriviaQA & 150 & 0.2 & 16 & 2000 \\
        *NLI & 300 & 0.2 & 64 & 2000 \\
        \bottomrule
    \end{tabular}
    \caption{Training hyper-parameters for our models. For optimization we employed ADAM with a learning rate of $10^{-3}$ which was halved when performance dropped between checkpoint (ckpt) intervals. We use $300$-dimensional word-embeddings from GloVe \citep{Pennington2014} as pre-trained word embeddings in all experiments. For regularization we make use of dropout on the computed non-contextual word representations $\mathbf{e}_w$ defined in \S\ref{sec:noncontextual_embeddings} with the same dropout mask for all words in a batch. For QA we additionally applied dropout on the projections computed in Eq.~\ref{eq:qa_proj}.
}\label{tab:training_details}
\end{table}

\subsection{Recognizing Textual Entailment}\label{sec:rte_impl}

\paragraph{Encoding} Analogous to DQA we encode our input sequences by BiLSTMs, however, for RTE we use conditional encoding \citep{Rocktschel2015} instead. Therefore, we initially process the embedded hypothesis $\mathbf{Q}$ by a BiLSTM and use the respective end states of the forward and backward LSTM as initial states for the forward and backward LSTM that processes the embedded premise $\mathbf{P}$. 

\paragraph{Prediction} We concatenate the outputs of the forward and backward LSTMs processing the premise $\boldsymbol{p}$, i.e., $\left[\tilde{\mathbf{p}}^{fw}_t; \tilde{\mathbf{p}}^{bw}_t \right] \in \mathbb{R}^{2n}$ and run each of the resulting $L_P$ outputs through a fully-connected layer with $\operatorname{ReLU}$ activation ($\mathbf{h}_t$) followed by a $\max$-pooling operation over time resulting in a hidden state $\mathbf{h} \in \mathbb{R}^{n}$. Finally, $\mathbf{h}$ is used to predict the RTE label as follows:

\begin{align}
    &\mathbf{h}_t = \operatorname{ReLU} \left(\operatorname{FC}\left( \begin{bmatrix} \tilde{\mathbf{p}}^{fw}_{t} \\ \tilde{\mathbf{p}}^{bw}_{t}  \end{bmatrix} \right) \right) \nonumber \\
    &\mathbf{h} = \underset{t}{\operatorname{maxpool}}\,\, \mathbf{h}_t \nonumber \\
    & p(c) = \operatorname{softmax}_c(\mathbf{v}_c \mathbf{h}) \quad , \, \mathbf{v}_c \in \mathbb{R}^n \label{eq:rte_prediction}
\end{align}
\noindent The probability of choosing category $c \in$ \{entailment, contradiction, neutral\} is defined in Eq.~\ref{eq:rte_prediction}. Finally, the model is trained to maximize the log-likelihood of the correct category label given probability distribution $p$.

\section{Reducing Training Data \& Dimensionality of Pre-trained Word Embeddings}
\label{sec:reduction_experiments}
We find that there is only little impact when using external knowledge on the RTE task when using a more sophisticated task model such as ESIM. We hypothesize that the attention mechanisms within ESIM together with powerful, pre-trained word representations allow for the recovery of some important lexical relations when trained on a large dataset. It follows that by reducing the number of training data and impoverishing pre-trained word representations the impact of using external knowledge should become larger.

To test this hypothesis, we gradually impoverish pre-trained word embeddings by reducing their dimensionality with PCA while reducing the number of training instances at the same time.\footnote{Although reducing either embedding dimensionality or data individually exhibit similar (but less pronounced) results we only report the joint reduction results here.} Our joint data and dimensionality reduction results are presented in Table~\ref{tab:data_reduction}. They show that there is indeed a slightly larger benefit when employing background knowledge from ConcepNet ($\mathcal{A}$) in the more impoverished settings with largest improvements when using around 10k examples and reduced dimensionality to 10. However, we observe that the biggest overall impact over the baseline ESIM model stems from our contextual refinement strategy (i.e., reading only the premise $\boldsymbol{p}$  and hypothesis $\boldsymbol{q}$) which is especially pronounced for the 1k and 3k experiments. This highlights once more the usefulness of our refinement strategy even without the use of additional knowledge. 

\begin{table*}[t]
    \small
    \centering
        \begin{tabular}{l c c c c c c}
            \toprule
            Dim/Data & 1/1k & 3/3k & 10/10k & 30/30k & 100/100k & 300/Full \\ 
            \midrule
            ESIM & 44.3 & 50.0 & 55.5 & 61.9 & 68.1 & 76.9 \\
            + $\boldsymbol{p}$ + $\boldsymbol{q}$ & 51.8(+7.5) & 55.8(+5.8) & 60.1(+4.6) & 65.0(+3.1) & 70.7(+2.6) & 78.1(+1.2) \\
            + $\boldsymbol{p}$ + $\boldsymbol{q}$ + $\mathcal{A}$ & 52.4(+0.6) & 57.9(+2.1) & 62.4(+2.3) & 66.6(+1.6) & 71.3(+0.6) & 78.8(+0.7) \\
            \bottomrule
        \end{tabular}    
        \caption{Development set results for MultiNLI (Matched + Mismatched) when reducing training data and embedding dimensionality with PCA. In parenthesis we report the relative differences to the respective result directly above.}\label{tab:data_reduction}
\end{table*}

\section{Further Analysis of Knowledge Utilization in RTE}
\label{sec:additional_analysis}

\begin{figure*}[ht!]
\begin{subfigure}{0.32\textwidth}
\includegraphics[width=\textwidth]{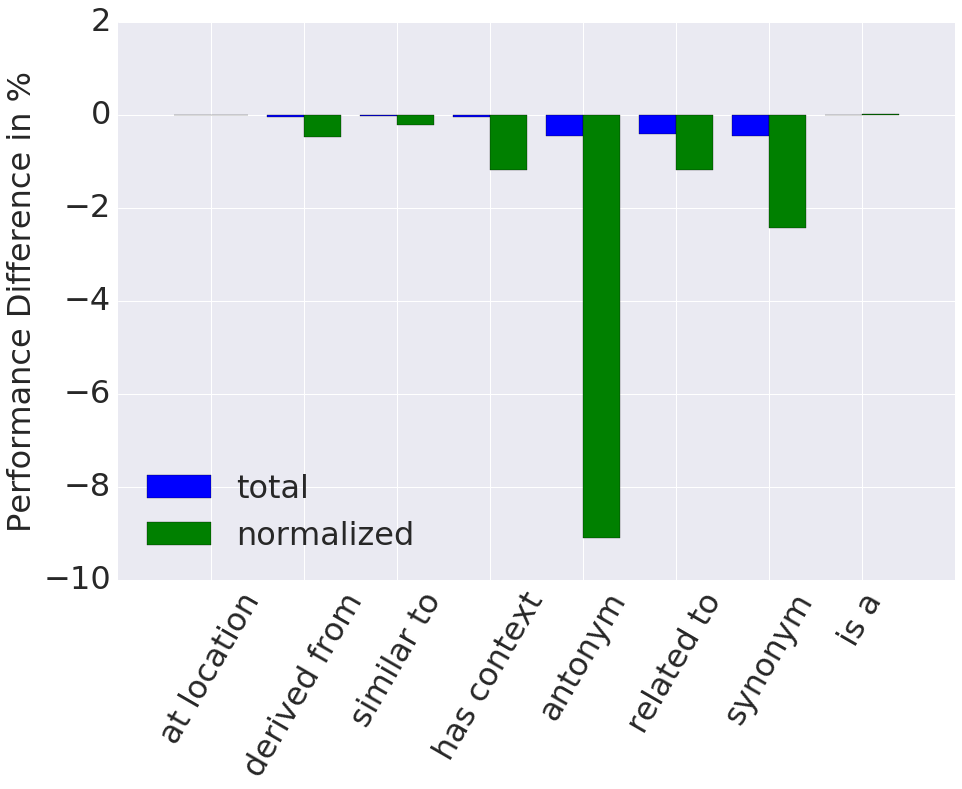}
\caption{BiLSTM on MultiNLI.}
\end{subfigure}~
\begin{subfigure}{0.3\textwidth}
\includegraphics[width=\textwidth]{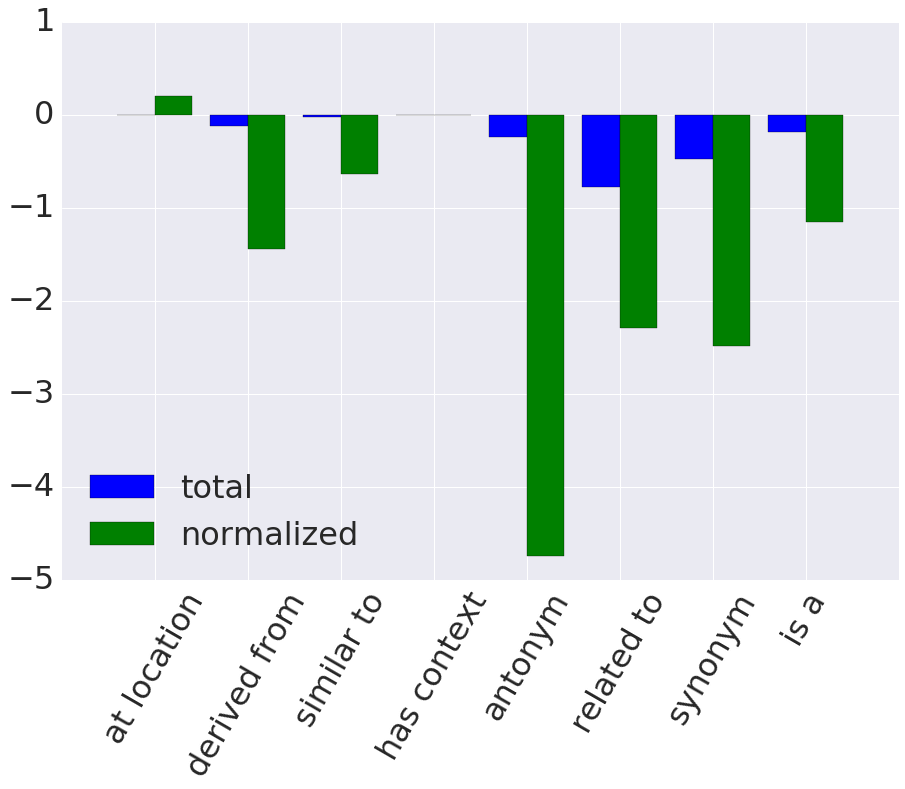}
\caption{ESIM on MultiNLI.}
\end{subfigure}~
\begin{subfigure}{0.3\textwidth}
\includegraphics[width=\textwidth]{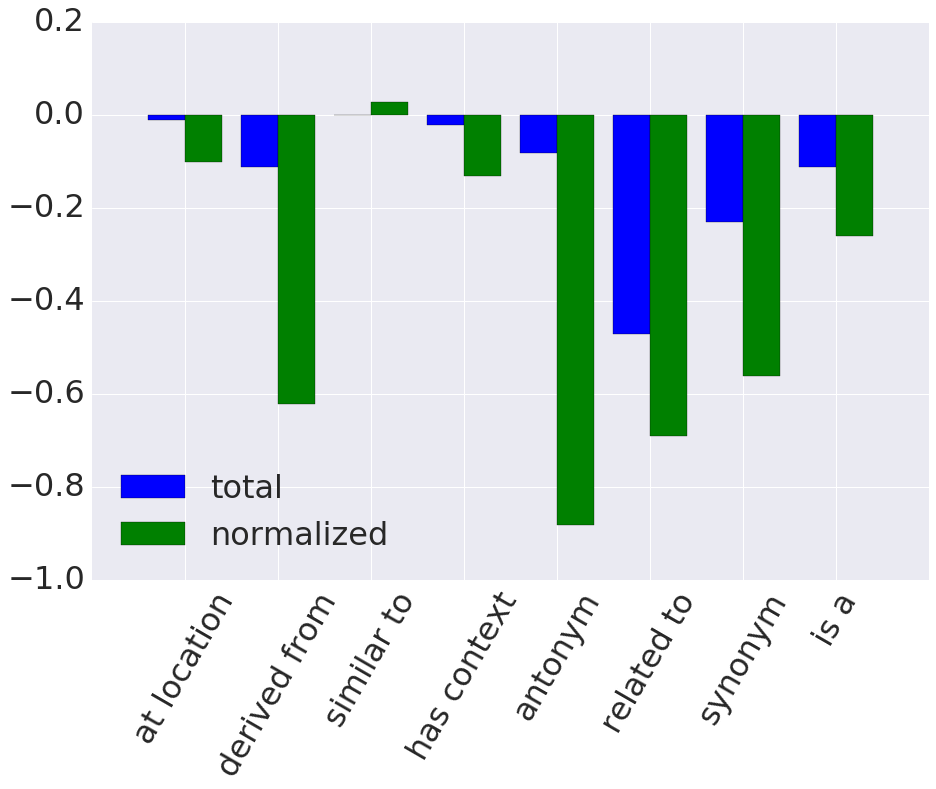}
\caption{BiLSTM on SQuAD.}
\end{subfigure}
\caption{Performance differences when ignoring certain types of knowledge, i.e., relation predicates during evaluation. Normalized performance differences are measured on the subset of examples for which an assertion of the respective relation predicate occurs.}\label{fig:ignore_relation}
\end{figure*}

\paragraph{Is additional knowledge used?} To verify whether and how our models make use of additional knowledge, we conducted several experiments. First, we evaluated models trained with knowledge on our tasks while not providing any knowledge at test time. This ablation drops performance by 3.7--3.9\% accuracy on MultiNLI, and by 4\% F1 on SQuAD. This indicates the model is refining the representations using the provided assertions in a useful way.

\paragraph{What knowledge is used?} After establishing that our models are somehow sensitive to semantics we wanted to find out which type of knowledge is important for which task. For this analysis we exclude assertions including the most prominent predicates in our knowledge base individually when evaluating our models. The results are presented in Figure~\ref{fig:ignore_relation}. They demonstrate that the biggest performance drop in total (blue bars) stems from \textit{related to} assertions. This very prominent predicate appears much more frequently than other assertions and helps connecting related parts of the 2 input sequences with each other. We believe that \textit{related to} assertions offer benefits mainly from a modeling perspective by strongly connecting the input sequences with each other and thus bridging long-range dependencies similar to attention. Looking at the relative drops obtained by normalizing the performance differences on the actually affected examples (green) we find that our models depend highly on the presence of \textit{antonym} and \textit{synonym} assertions for all tasks as well as partially on \textit{is a} and \textit{derived from} assertions. This is an interesting finding which shows that the sensitivity of our models is selective wrt. the type of knowledge and task. The fact that the largest relative impact stems from \textit{antonyms} is very interesting because it is known that such information is hard to capture with distributional semantics contained in pre-trained word embeddings.


\end{document}